\documentclass[letterpaper, 10 pt, conference]{ieeeconf}  

\IEEEoverridecommandlockouts                              

\usepackage{amsmath,amsfonts}
\usepackage{array}
\usepackage[caption=false,font=footnotesize,labelfont=rm,textfont=rm]{subfig}
\usepackage{textcomp}
\usepackage{stfloats}
\usepackage{url}
\usepackage{verbatim}
\usepackage{graphicx}
\usepackage{cite}
\usepackage{times}
\usepackage{epsfig}
\usepackage{float}
\usepackage{wrapfig}
\usepackage{algorithm, algorithmic}
\usepackage{bm,xspace}
\usepackage{comment}
\usepackage{multirow}
\usepackage{threeparttable}
\usepackage{balance}
\usepackage{booktabs}
\usepackage{etoolbox,siunitx}
\usepackage{calc}
\usepackage{pifont,hologo}
\usepackage{makecell}
\usepackage{adjustbox}
\usepackage{float}
\usepackage[colorlinks=black,citecolor=green]{hyperref}
\makeatletter
\newenvironment{breakablealgorithm}
{
		\begin{center}
			\refstepcounter{algorithm}
			\hrule height.8pt depth0pt \kern2pt
			\renewcommand{\caption}[2][\relax]{
				{\raggedright\textbf{\ALG@name~\thealgorithm} ##2\par}%
				\ifx\relax##1\relax 
				\addcontentsline{loa}{algorithm}{\protect\numberline{\thealgorithm}##2}%
				\else 
				\addcontentsline{loa}{algorithm}{\protect\numberline{\thealgorithm}##1}%
				\fi
				\kern2pt\hrule\kern2pt
			}
		}{
		\kern2pt\hrule\relax
	\end{center}
}
\makeatother

\title{\LARGE \bf
Block-Map-Based Localization in Large-Scale Environment
}

\author{Yixiao Feng$^{1,2*}$, Zhou Jiang$^{1,3*}$, Yongliang Shi$^{4\dag}$, Yunlong Feng$^{5}$, Xiangyu Chen$^{1}$, \\ Hao Zhao$^{1}$, Guyue Zhou$^{1}$
\thanks{* Equal Contribution, $^{1}$Institute for AI Industry Research (AIR), Tsinghua University; $^{2}$University of New South Wales; $^{3}$Beijing Institute of Technology; $^{4}$QiYuan Lab; $^{5}$ShanghaiTech University.}%
\thanks{$\dag$ Corresponding author. shiyongliang@qiyuanlab.com}%
\thanks{Sponsored by Tsinghua-Toyota Joint Research Fund (20223930097).}
}

\begin{document}

\maketitle
\thispagestyle{empty}
\pagestyle{empty}

\begin{abstract}
Accurate localization is an essential technology for the flexible navigation of robots in large-scale environments. Both SLAM-based and map-based localization will increase the computing load due to the increase in map size, which will affect downstream tasks such as robot navigation and services. To this end, we propose a localization system based on Block Maps (BMs) to reduce the computational load caused by maintaining large-scale maps.
Firstly, we introduce a method for generating block maps and the corresponding switching strategies, ensuring that the robot can estimate the state in large-scale environments by loading local map information. 
Secondly, global localization according to Branch-and-Bound Search (BBS) in the 3D map is introduced to provide the initial pose. 
Finally, a graph-based optimization method is adopted with a dynamic sliding window that determines what factors are being marginalized whether a robot is exposed to a BM or switching to another one, which maintains the accuracy and efficiency of pose tracking.
Comparison experiments are performed on publicly available large-scale datasets. 
Results show that the proposed method can track the robot pose even though the map scale reaches more than 6 kilometers, while efficient and accurate localization is still guaranteed on NCLT\cite{ncarlevaris} and M2DGR\cite{9664374}.
Codes and data will be publicly available on \href{https://github.com/YixFeng/block_localization.git}{https://github.com/YixFeng/block\_localization}.
\end{abstract}
\section{INTRODUCTION}
For large-scale robotic automation in GPS-denied environments, such as indoor industrial environments, and underground mining, efficient and precise localization is a fundamental capability required by most autonomous mobile systems. 

GNSS/INS systems suffer from the signal block in urban scenarios\cite{gu2015gnss}, which makes the localization result unreliable. 
The maturation of the 3D map construction achieved through the offline processing of LiDAR or camera data is evident. 
Particularly prominent within the domain of robotic systems operating within predetermined memory limitation, the employment of LiDAR/visual odometry for the purpose of localization introduces a salient challenge arising from the proclivity for error accumulation.
This characteristic engenders deviations in the proficient execution of tasks entrusted to robotic entities.
In light of this, an alternate strategy of heightened efficacy emerges, characterized by the utilization of the meticulously organized map, engendered via offline methodologies, as a steadfast and dependable modality of localization. 
Consequently, the adoption of map-based localization assumes paramount importance for autonomous mobile systems with distinct operational objectives such as remote delivery and mobile operations, among others.
Using dense maps \cite{chen2021range} is usually more accurate but suffers from high computational time. The use of feature maps\cite{jo2015precise} decreases the computational time and makes the localization task more suitable for real-time application. 
Even so, the increase in the scale of the map aggravates the performance overhead and the search domain uncertainty of the registration process, which leads to a decrease in the real-time and accuracy of robot localization. 

Various approaches based on camera\cite{xu20173d} and LiDAR\cite{caballero2021dll} are proposed for Map-based localization. During which, LiDAR has been widely used for environment perception thanks to their accurate range measurements. Besides, as a result of accurate short-term motion constraints at high frequency, IMU is adopted for compensating deficiencies of LiDAR. Popular solutions for Lidar-Inertial based state estimation can be divided into two categories: filtering-based and optimization-based approaches. Filtering-based approaches infer the most likely state from available measurements and uncertainties, while optimization-based approaches try to minimize reprojection error to find the optimal states. However, the filter-based algorithm assumes Markovianity, which is a fundamental constraint limiting its performance, and global batch optimization cannot guarantee real-time performance\cite{barfoot2017state}.

\begin{figure}[!t]
\vspace{0.1in}
\centering
\includegraphics[width=3.4in]{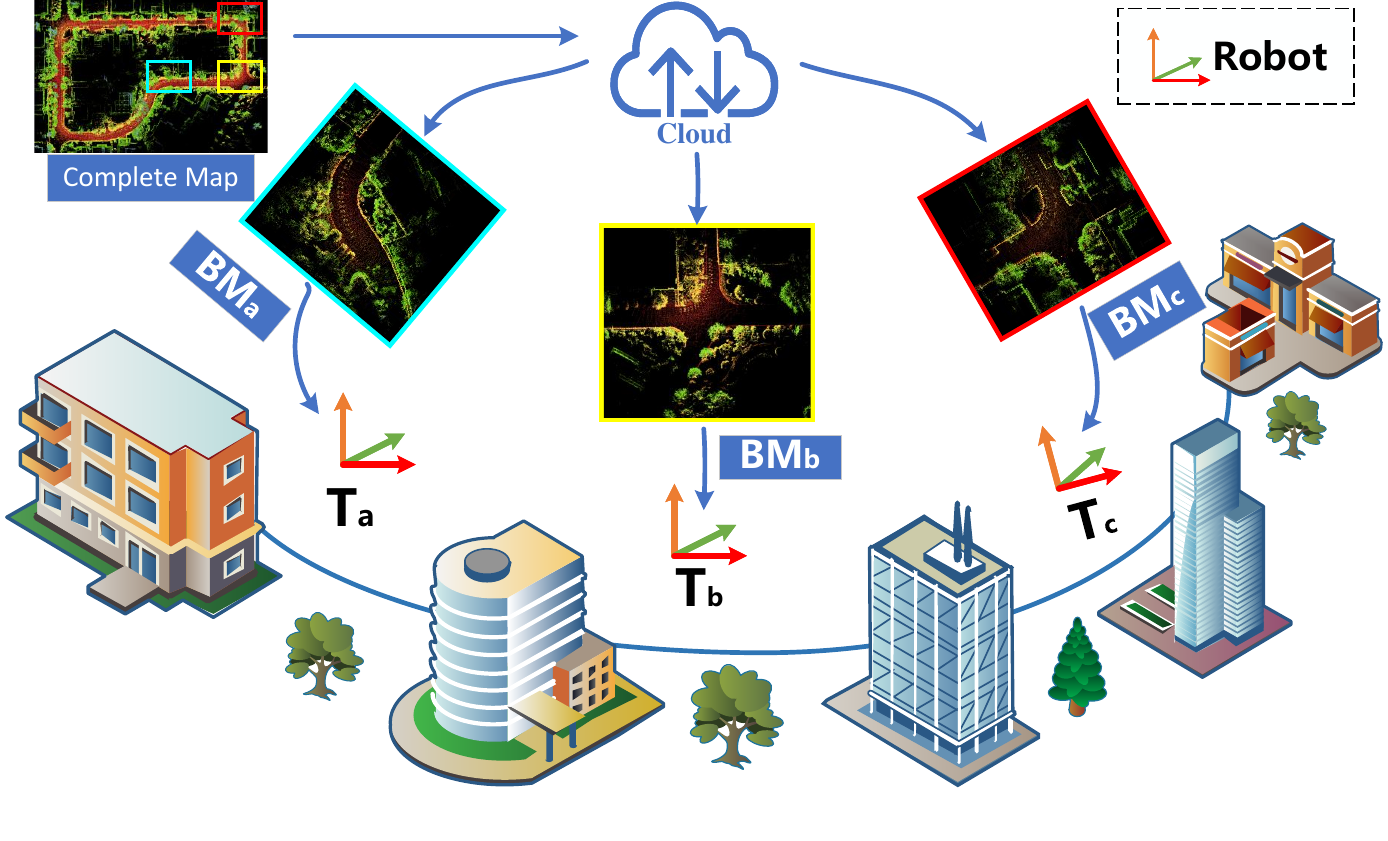}
\caption{Dividing a large-scale map into multiple block maps has the potential to enable robots to utilize limited resources to achieve arbitrary scale navigation and service tasks. Consequently, the utilization of block map-based localization becomes indispensable.}
\label{teaser}
\vspace{-0.2in}
\end{figure}
To address these limitations, we propose a BM generation and maintenance method and the corresponding BM-based localization system \ref{teaser}. We employ keyframe stitching instead of segmenting the entire global map. This approach ensures spatial continuity between block maps through the overlap of keyframes, thereby preventing the loss of correlation information between the robot's laser point cloud and the map when transiting between maps. 
Consequently, given a coarse pose of the robot, BBS is utilized on the pyramid of a given BM to get the initial pose. On top of this, for the trade-off of accuracy and efficiency in pose tracking, we propose a factor graph-based optimization method with the dynamic sliding window that maintains different factors in cases of the same BM and switching BM. 

In summary, our contributions are as follows:
\begin{itemize}
\item A BM-based localization system in a large-scale environment is proposed for the first time.
\item A BM generation method and corresponding switching strategy is proposed which maintains the spatial continuity of adjacent BMs.
\item A factor graph-based optimization method with the dynamic sliding window based on BMs is proposed to achieve accurate and reliable state estimation. 
\item We achieve the best performance on publicly available large-scale datasets. 
\end{itemize}

\section{Related Work}
Building a 3D map of an unknown environment is crucial for map-based localization, which provides necessary information for path planning. Due to the dense and accurate depth measurements of environments, 3D LiDAR has emerged as an essential sensor for robots. 
The cartographer\cite{hess2016real} make the laser fan project onto the horizontal plane using an IMU, which is applicable to relatively flat ground scenes.
LOAM\cite{zhang2014loam} introduces planes and edges as features to achieve low-drift and low-computational complexity. Inspired by LOAM, there have been multiple variations of LOAM that enhance its performance. By incorporating IMU data through tight coupling for the state estimation\cite{xu2022fast,li2021towards,qin2020lins}, SLAM systems naturally become more precise and flexible. 
Iterative closest points (ICP), Normal Distribution Transform (NDT) and their variances are also utilized as state estimation methods for odometry of LiDAR SLAM\cite{ct-icp,LiTAMIN2,ndtmapping}. NDT reduces the computational time with respect to ICP approaches, while keeping the accuracy. None of the above methods implements the solution of dividing a large-scale map into several block maps.

Given the map generated by LiDAR SLAM, we often distinguish localization problem between global localization and pose tracking \cite{2005Probabilistic}.
For global localization, Monte Carlo localization\cite{thrun2001robust} is a popular framework, which uses a particle filter to estimate the robot’s pose and is widely used in robot localization systems\cite{chen2021range,9196708}. This filtering methodology should be more robust to local minima because the particles should ideally come to a consensus through additional measurements — though this is dependent on random sampling and can make no time-based optimality guarantees. 
As one of the most successful algorithms for global optimization problem\cite{breuel2003implementation,hartley2009global}, BBS has been widely used for localization in autonomous driving\cite{chen2021pole,wolcott2015fast}.

High-rate IMU measurements can effectively compensate for the motion distortion in a LiDAR scan. To achieve robust state estimation in challenging situations, LiDAR fused with IMU has become prevailing in the process of map-based pose tracking. 
Levinson\cite{levinson2007map} adopt the particle filter to combine multiple observations for state estimation.
Zhen\cite{zhen2017robust} uses the Error State Kalman Filter (ESKF)  for sensor fusion and is combined with a Gaussian Particle Filter (GPF) for measurements update. Iterated Error State Kalman Filter (IESKF) \cite{qin2020lins} is designed to correct the estimated state recursively by generating new feature correspondences in each iteration, which enables robust and efficient navigation for ground vehicles in feature-less scenes.  
Lie Group has been introduced to define the state and reduce the linearization error in Invariant Extended Kalman Filter (Invariant EKF)\cite{9870841}, Iterative Error State Kalman Filter\cite{tao2022adaptive} and Invariant Unscented Kalman Filter (Invariant UKF)\cite{brossard2017unscented}. Nevertheless, the Markov assumptions limit the performance of filter-based state estimation, while the factor graph optimization (FGO) presents a smoothing state estimation framework for flexible sensor fusion\cite{nubert2022graph,ding2020lidar}. As the FGO with high accuracy is computationally expensive, localization with sliding window factor graphs\cite{ding2020lidar,wilbers2019approximating} are proposed to provide highly accurate pose estimates in real-time. 


\section{Method}
\begin{figure*}[ht]
\centering
\includegraphics[width=7.0in]{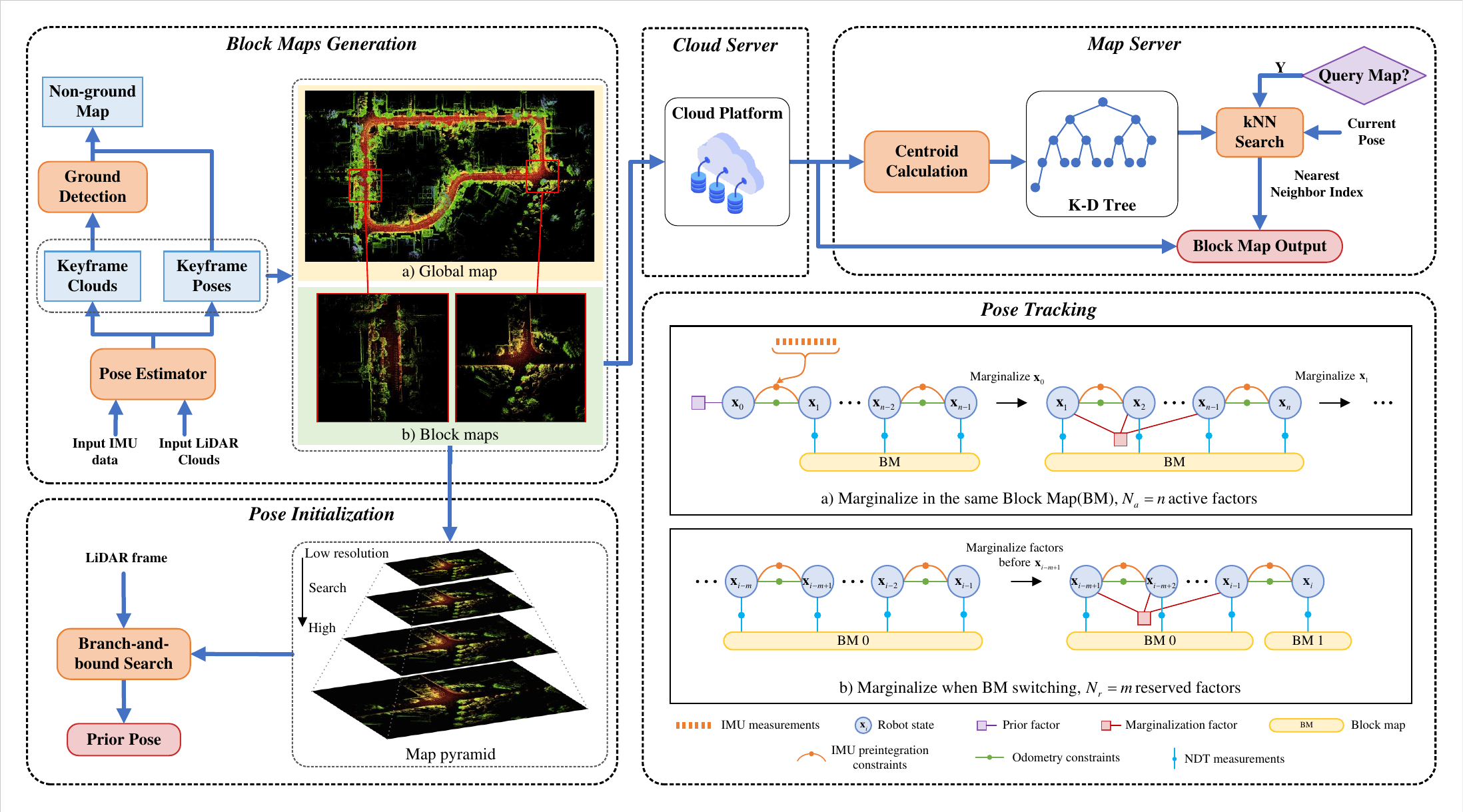}
\caption{Overview of our Block-Map-Based Localization which consists of four main modules: a block maps generation module, a pose initialization module, a map retrieval server module and a pose tracking module.}
\label{fig_sim}
\end{figure*}

\subsection{System Overview}
As Fig.\ref{fig_sim} shows, based on the poses $\mathbf{T}^{W_{\textrm{off}}}_{\textrm{off},m}$ obtained from offline graph-based SLAM\cite{nclt}, we opt for keyframe fusion as opposed to segmenting the entire global map. This approach ensures spatial continuity between block maps by leveraging the overlap among keyframes, thereby preventing the loss of correlation information between the robot's laser point cloud and the map during map transitions. Additionally, we calculate the map centroid and establish a KD-Tree structure to maintain the map retrieval repository. Subsequently, we emulate cloud storage using ROS Services to store the maps in the cloud. The retrieval through BBS on three-dimensional maps at different resolutions enables the robot to achieve coarse-to-fine global localization. To ensure robustness in the robot's pose tracking for localization, we employ sliding window-based factor graph optimization to ensure that the robot leverages a sufficient amount of historical information. When switching BMs, due to the loss of a substantial amount of prior map information, we dynamically adjust the sliding window, and aim to mitigate the interference caused by irrelevant historical data.

\subsection{Block Maps Generation}
\begin{figure}[t]
    \centering
    \subfloat[Ours]{
        \includegraphics[height=0.25\linewidth]{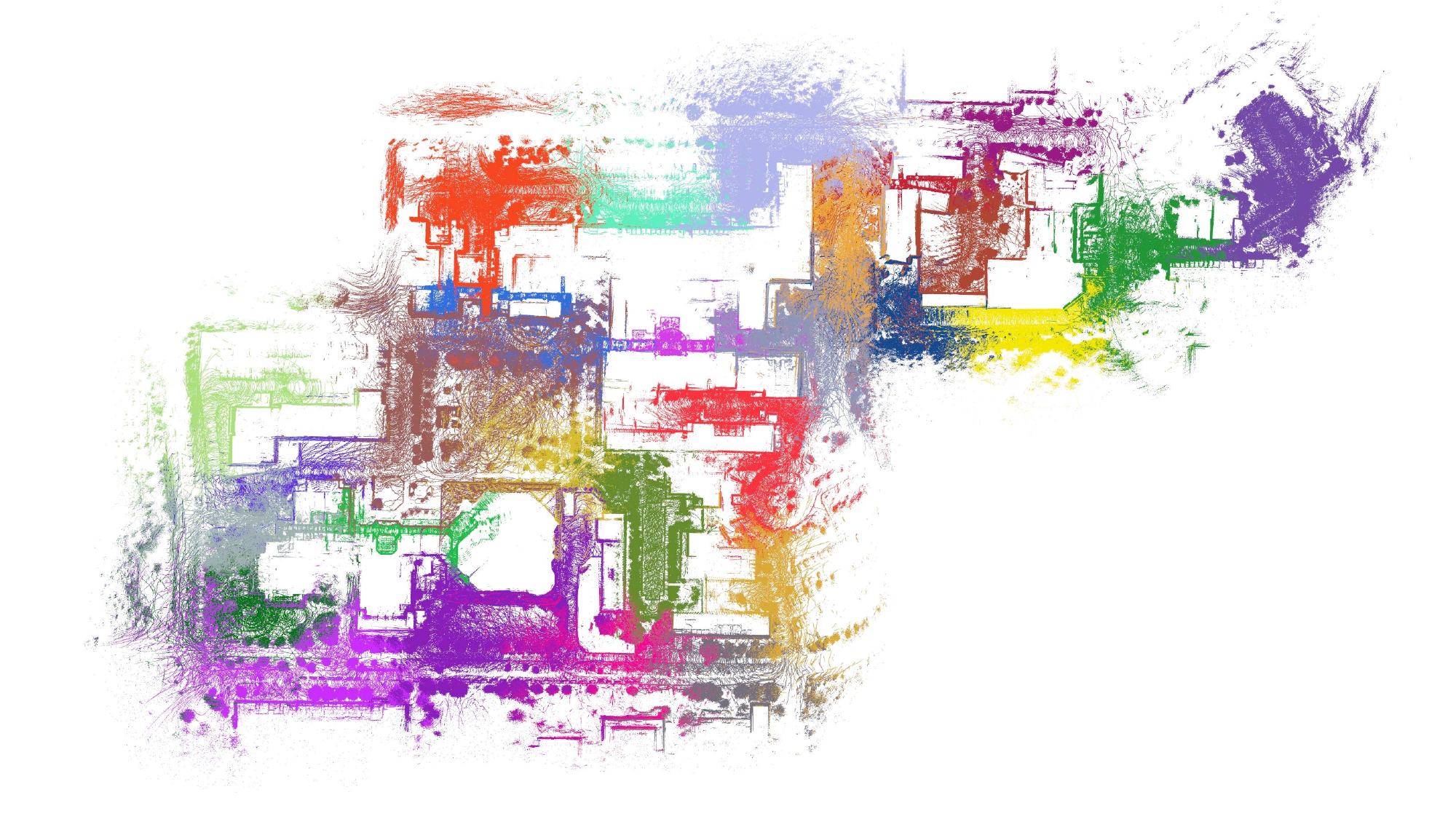}
        \label{fig:bms_ours}
    }
    \subfloat[Gao's\cite{gaoxiang}]{
        \includegraphics[height=0.25\linewidth]{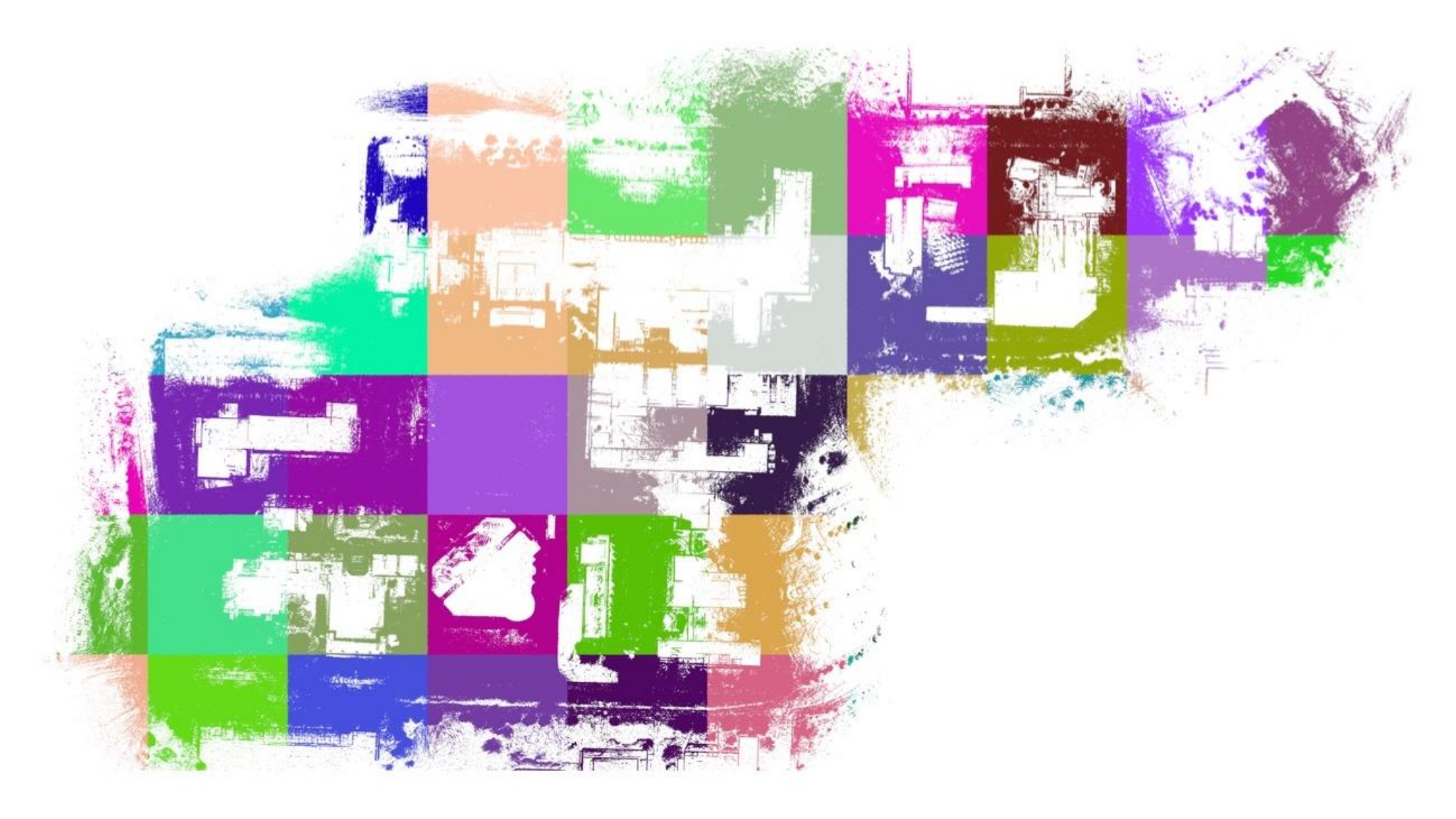}
        \label{fig:bms_gaos}
    }
    \caption{The result comparison of different block maps separation methods. Our method considers the overlap between adjacent block maps, and the robot will not lose prior information when switching maps at the boundary.}
    \label{fig:BMs_comparison}
\end{figure}

In the Block-Map-Based localization method, maintaining spatial continuity in robotic observations during BM transitions is of paramount importance to ensure smooth estimates and stability. To address this, our proposed methodology focuses on the deployment of keyframe-based integration to build BMs, instead of the more traditional method of partitioning a comprehensive map into isolated BMs, as Fig.\ref{fig:bms_gaos}, which neglects the relational integrity of map information between contiguous BMs. The intricate procedure is elucidated in Algorithm \ref{alg:bm-generation}.

\begin{breakablealgorithm}
\renewcommand{\algorithmicrequire}{\textbf{Input:}}
\renewcommand{\algorithmicensure}{\textbf{Output:}}
\caption{Block Maps (BMs) Generation Method}
\label{alg:bm-generation}
\begin{algorithmic}[1]
        \REQUIRE LiDAR raw points in all $k$ scans $\{^{L}\mathbf{p}_i\}_k$; \par
        \hspace{1.4em}Highly precise poses obtained from offline SLAM $\mathbf{T}^{W_{\textrm{off}}}_{\textrm{off}, m}$; \par
        \hspace{1.4em}Extrinsic parameters $\mathbf{T}^{L}_{\textrm{off}}$;\par
        \hspace{1.4em}Block map size $\mathcal{S}$.
        \ENSURE The block maps $\{^{W_{l}}\mathbf{BM}_j\}$.
        \STATE {Block maps count $\mathcal{C}=0$;}
        \FORALL{pointcloud $^{L}\mathbf{p}_i$ in $\{^{L}\mathbf{p}_i\}_k$}
            \STATE Find the closest pose $\mathbf{T}^{W_{\textrm{off}}}_{\textrm{off}}$ to the pointcloud timestamp;
            \STATE Transform LiDAR points via $^{\textrm{off}}\mathbf{p}_i = ({\mathbf{T}^{L}_{\textrm{off}}})^{-1} {^{L}\mathbf{p}_i}$; \par
            \IF{it is the first pointcloud}
                \STATE Record the translation part of the corresponding pose as $\widehat{\mathbf{t}}^{W_{\textrm{off}}}_{\textrm{off}}$;
                \STATE Create a new $^{W_l}\widehat{\mathbf{BM}}$ for temporary storage; 
            \ENDIF
            \STATE $\mathbf{t}^{W_{\textrm{off}}}_{\textrm{off}}$ denotes the translation part of $\mathbf{T}^{W_{\textrm{off}}}_{\textrm{off}}$;
            \IF{$Euclidean\_dist(\widehat{\mathbf{t}}^{W_{\textrm{off}}}_{\textrm{off}}, \mathbf{t}^{W_{\textrm{off}}}_{\textrm{off}}) \leq \mathcal{S}$}
                \STATE $^{W_l}\widehat{\mathbf{BM}}={^{W_l}\widehat{\mathbf{BM}}}\cup{\mathbf{T}^{L}_{\textrm{off}}}{\mathbf{T}^{W_{\textrm{off}}}_{\textrm{off}}}{^{\textrm{off}}\mathbf{p}_i}$;
            \ELSE
                \STATE $\widehat{\mathbf{t}}^{W_{\textrm{off}}}_{\textrm{off}} = {\mathbf{t}}^{W_{\textrm{off}}}_{\textrm{off}}$;
                \IF{$\mathcal{C}$ is equal to $0$ or $1$}
                    \STATE Store $^{W_l}\widehat{\mathbf{BM}}$ with its count as $^{W_l}\mathbf{BM}_j$;
                    \STATE Compute its centroid point $^{W_l}\mathbf{p}_c$ and build a KD-Tree using all centroids;
                    \STATE $\mathcal{C}=\mathcal{C}+1$; 
                \ELSE
                    \STATE Compute the centroid of $^{W_l}\widehat{\mathbf{BM}}$;
                    \STATE Find its nearest neighbor in KD-Tree and save the minimum distance $\mathbf{d}$ and index $\mathbf{n}$;
                    \IF{$\mathbf{d} \geq 0.5\mathcal{S}$}
                        \STATE Redo lines 15 to 17;
                    \ELSIF{$0.1\mathcal{S} < \mathbf{d} < 0.5\mathcal{S}$}
                        \STATE $^{W_l}\mathbf{BM}_\mathbf{n} = {^{W_l}\mathbf{BM}_\mathbf{n}} \cup {^{W_l}\widehat{\mathbf{BM}}}$;
                        \STATE Renew the centroid and update the KD-Tree;
                    \ENDIF
                \ENDIF
                \STATE Clear the $^{W_l}\widehat{\mathbf{BM}}$.
            \ENDIF
        \ENDFOR
\end{algorithmic}
\end{breakablealgorithm}
In our approach, as the robot frame in offline SLAM denoted as $^{\textrm{off}}$ is different from the lidar frame, we leverage the extrinsics $\mathbf{T}^{L}_{\textrm{off}}$ to reproject the points into the frame $^{\textrm{off}}$ and build the BMs by $\mathbf{T}^{W_{\textrm{off}}}_{\textrm{off}}$. To filter out excessively overlapping BMs while remaining more map information, we compute the centroids of the BMs to maintain a KD-Tree. When a new candidate BM is completed, we use its centroid to search for the nearest BM, and if the distance $\mathbf{d}$ satisfies $0.1\mathcal{S} < \mathbf{d} < 0.5\mathcal{S}$, we merge these two BMs to maintain an appropriate map overlap.
\par This strategy effectively capitalizes on the overlapping segments among keyframes, thereby ensuring a seamless scan-to-map alignment during the map transition process.

\subsection{Pose Initialization}

Initializing the pose of robots is the beginning of the whole system. The initial pose gives the first prior for the factor graph to stabilize the non-linear optimization process. Inspired by \cite{carto}, we build up a pyramid of different resolution maps for searching. In the searching step, we start from the minimum resolution map and recursively iterate from the highest score branch to the lowest score branch. To reduce the computing cost of BBS, we simplified the search process by giving an adapted initial score $\mathbf{S}_0$.

We are interested in finding the accurate $\epsilon$ in the initial block map, as shown in the Eq.\eqref{eq:BBS}:
\begin{equation}
    \epsilon^* = \mathop{\arg\min}_{\epsilon \in \mathbb{L}} \text{M}(\mathbf{T}_\epsilon \mathbf{h}_k),
    \label{eq:BBS}
\end{equation}
where $\epsilon^*$ is the optimizing pose in the block maps. $\mathbf{T}_\epsilon$ is pose $\epsilon$ represented in Lie Group. $\text{M}(\cdot)$ is a measurement function to get the score of current pose $\epsilon$ in 2D block map planers just as \cite{carto}. 

Firstly, a map pyramid $\mathbf{P}_m$ from level $0$ to level $N$ is built, and each level is represented as $\mathbb{L}_i\in \mathbf{P}_m (i = 0, 1, \cdots, N)$. The resolution of $\mathbb{L}_i$ is half that of $\mathbb{L}_{i-1}$. Secondly, we choose the angular $\delta_\theta$ and the linear $r$ step sizes in searching. For each $\mathbb{L}_i (i = 0, 1, \cdots, N)$, create a search window $\mathcal{W}^i$ in $\mathbb{L}_i$.
\begin{equation}
    w^i_x = \left \lceil \frac{\mathbf{W}^i_x}{r} \right \rceil,  w^i_y = \left \lceil \frac{\mathbf{W}^i_y}{r} \right \rceil, w^i_\theta = \left \lceil \frac{\mathbf{W}^i_\theta}{\delta_\theta} \right \rceil,
\end{equation}

\begin{equation}
    \overline{\mathcal{W}}^i = \{ -w^i_x, \cdots, w^i_x \} \times \{ -w^i_y, \cdots, w^i_y \} \times \{ -w^i_\theta, \cdots, w^i_\theta \},
\end{equation}
\begin{equation}
    \mathcal{W}^i = \{ \epsilon_0 + (rj_x, rj_y, rj_\theta)\ |\ (rj_x, rj_y, rj_\theta)\in \overline{\mathcal{W}}^i\}.
\end{equation}
Finally, adaptively select a base score for branch and bound search.

\subsection{Pose Tracking}
\subsubsection{Problem Formulation}
We propose our graph-based optimization framework with a changeable sliding window size. In our case, we define a set of poses $\boldsymbol{x}=\{\boldsymbol{x}_{t_s},\cdots,\boldsymbol{x}_{t_e} \}$ and observations $\boldsymbol{z}$, where $t_s$ and $t_e$ are the first and last pose in the sliding window respectively. We seek to estimate the most likely states of the robot $\boldsymbol{x}$ over a set of measurements $\boldsymbol{z}$ from various sensors inside the current sliding window. This problem can be formulated as a maximum a posteriori (MAP) problem:
\begin{equation}
    \boldsymbol{x}^*=\mathop{\arg\max}_{\boldsymbol{x}}P(\boldsymbol{z}\ |\ \boldsymbol{x})P(\boldsymbol{x}),
    \label{eq:MAP}
\end{equation}
Assuming the noise follows the Gaussian distribution and is independent, Eq.\eqref{eq:MAP} can be rewritten as a sum of minimum cost functions:
\begin{equation}
    \boldsymbol{x}^* = \mathop{\arg\min}_{\boldsymbol{x}}\sum_k\boldsymbol{e}_k(\boldsymbol{x},\boldsymbol{z}_k)^\top\boldsymbol{\Omega}_k \boldsymbol{e}_k(\boldsymbol{x},\boldsymbol{z}_k),
    \label{eq:min_cost}
\end{equation}
$\boldsymbol{e}_k(\boldsymbol{x},\boldsymbol{z}_k)$ denotes the cost functions between robot state vector $\boldsymbol{x}$ and measurements $\boldsymbol{z}_k$, and $\boldsymbol{\Omega}_k$ is the corresponding information matrices. To be specific, we introduce three types of error functions and a prior from sliding window marginalization:
\begin{equation}
\begin{aligned}
    \boldsymbol{x}^* = \mathop{\arg\min}_{\boldsymbol{x}}&\sum_k\boldsymbol{e}^{\textrm{odm}}(\boldsymbol{x},\boldsymbol{z}_k^{\textrm{odm}})^\top \boldsymbol{\Omega}_k^{\textrm{odm}}\boldsymbol{e}^{\textrm{odm}}(\boldsymbol{x},\boldsymbol{z}_k^{\textrm{odm}}) \\
    +&\sum_k\boldsymbol{e}^{\textrm{map}}(\boldsymbol{x},\boldsymbol{z}_k^{\textrm{map}})^\top \boldsymbol{\Omega}_k^{\textrm{map}}\boldsymbol{e}^{\textrm{map}}(\boldsymbol{x},\boldsymbol{z}_k^{\textrm{map}}) \\
    +&\sum_k\boldsymbol{e}^{\textrm{imu}}(\boldsymbol{x},\boldsymbol{z}_k^{\textrm{imu}})^\top \boldsymbol{\Omega}_k^{\textrm{imu}}\boldsymbol{e}^{\textrm{imu}}(\boldsymbol{x},\boldsymbol{z}_k^{\textrm{imu}}) \\
    +&\mathcal{F}^{\textrm{marg}}(\boldsymbol{x}).
\end{aligned}
\end{equation}
We differentiate between error functions by superscripts, $^{\textrm{map}}$ for NDT matching errors, $^{\textrm{odm}}$ for odometry errors, and $^{\textrm{imu}}$ for IMU preintegration measurements between subsequent poses. Additionally, $\mathcal{F}^{\textrm{marg}}(\boldsymbol{x})$ stands for the prior information that derives from marginalized measurements outside the sliding window.

\subsubsection{Point Deskewing}
Due to the particular mechanism of the rotating 3D lidar, the raw scan received, $\mathcal{P}_k$, is distorted when moving at a fast speed. To cope with this issue, we use the steady-speed motion model from previous poses to predict lidar translation and obtain the rotation from IMU forward propagation. Denote the lidar pose of the k-th sweep as $\mathbf{T}_k$ and the transformation between two consecutive frames $k-1$ to $k$ can be represented by: 
\begin{equation}
    \xi^{k-1}_k=\log(\mathbf{T}_{k-2}^{-1}\mathbf{T}_{k-1})\in \mathfrak{se}(3),
\end{equation}
Then, we use the timestamp $t\in(t_{k},t_{k'}]$ of each point to interpolate the lidar motion, where $t_{k}$ is the beginning of the k-th sweep and $t_{k'}$ is the end: 
\begin{equation}
    \mathbf{T}_k(t)=\mathbf{T}_{k-1}\exp(\frac{t-t_k}{t_{k'}-t_k}{\xi^{k-1}_k}),
\end{equation}
The raw points $\mathcal{P}_k$ can be corrected into the starting pose of the sweep: 
\begin{equation}
    \widetilde{\mathcal{P}}_k=\{\mathbf{T}_k(t)\mathbf{p}_k \ |\ \mathbf{p}_k\in\mathcal{P}_k\}.
\end{equation}

\subsubsection{Dynamic Sliding Window}
In this section, we introduce our marginalization strategy with changeable sliding window size. The number of active keyframes in the current sliding window depends on two BM-related cases shown in the pose tracking module of Fig.\ref{fig_sim}. In the first case, when the robot navigates in a single BM, we keep a maximum of $N_a=20$ keyframes and marginalize the old set of variables. By linearizing these factors, the linear system becomes:
\begin{equation}
    \begin{bmatrix}
        \mathbf{H}_{\alpha \alpha} & \mathbf{H}_{\alpha \beta} \\
        \mathbf{H}_{\beta \alpha} & \mathbf{H}_{\beta \beta}
    \end{bmatrix}
    \begin{bmatrix}
        \boldsymbol{x}_\alpha \\ \boldsymbol{x}_\beta
    \end{bmatrix} =
    \begin{bmatrix}
        \mathbf{b}_\alpha \\ \mathbf{b}_\beta
    \end{bmatrix},
\end{equation}
where $\beta$ denotes the set of variables that have to be marginalized, and all variables dependent on them are regarded as $\alpha$. Then, we apply the Schur complement, which yields a new linear system $\widehat{\mathbf{H}_{\alpha\alpha}}\boldsymbol{x}_\alpha=\widehat{\mathbf{b}_\alpha}$ with:
\begin{equation}
    \widehat{\mathbf{H}_{\alpha\alpha}}=\mathbf{H}_{\alpha\alpha}-\mathbf{H}_{\alpha\beta}\mathbf{H}_{\beta\beta}^{-1}\mathbf{H}_{\beta\alpha},
\end{equation}
\begin{equation}
    \widehat{\mathbf{b}_\alpha}=\mathbf{b}_\alpha-\mathbf{H}_{\alpha\beta}\mathbf{H}_{\beta\beta}^{-1}\mathbf{b}_\beta.
\end{equation}
This strategy creates a marginalization factor connecting all variables related to the factors removed.
\par In another case, the robot traverses the boundary of BMs which means the system has to retrieve a new BM for the source of scan-to-map alignment. Towards the goal of retaining enough but not redundant historical information, we only reserve $N_r = 5$ keyframes in the current sliding window when BM switching and all factors before will be marginalized as a prior. The design of reserved factors corresponds to the information involved in the overlapping zone of adjacent BMs. It aims to estimate a smooth trajectory while switching the map instead of violent fluctuations.

\begin{figure*}[t]
    \subfloat[]{\includegraphics[height=0.135\linewidth]{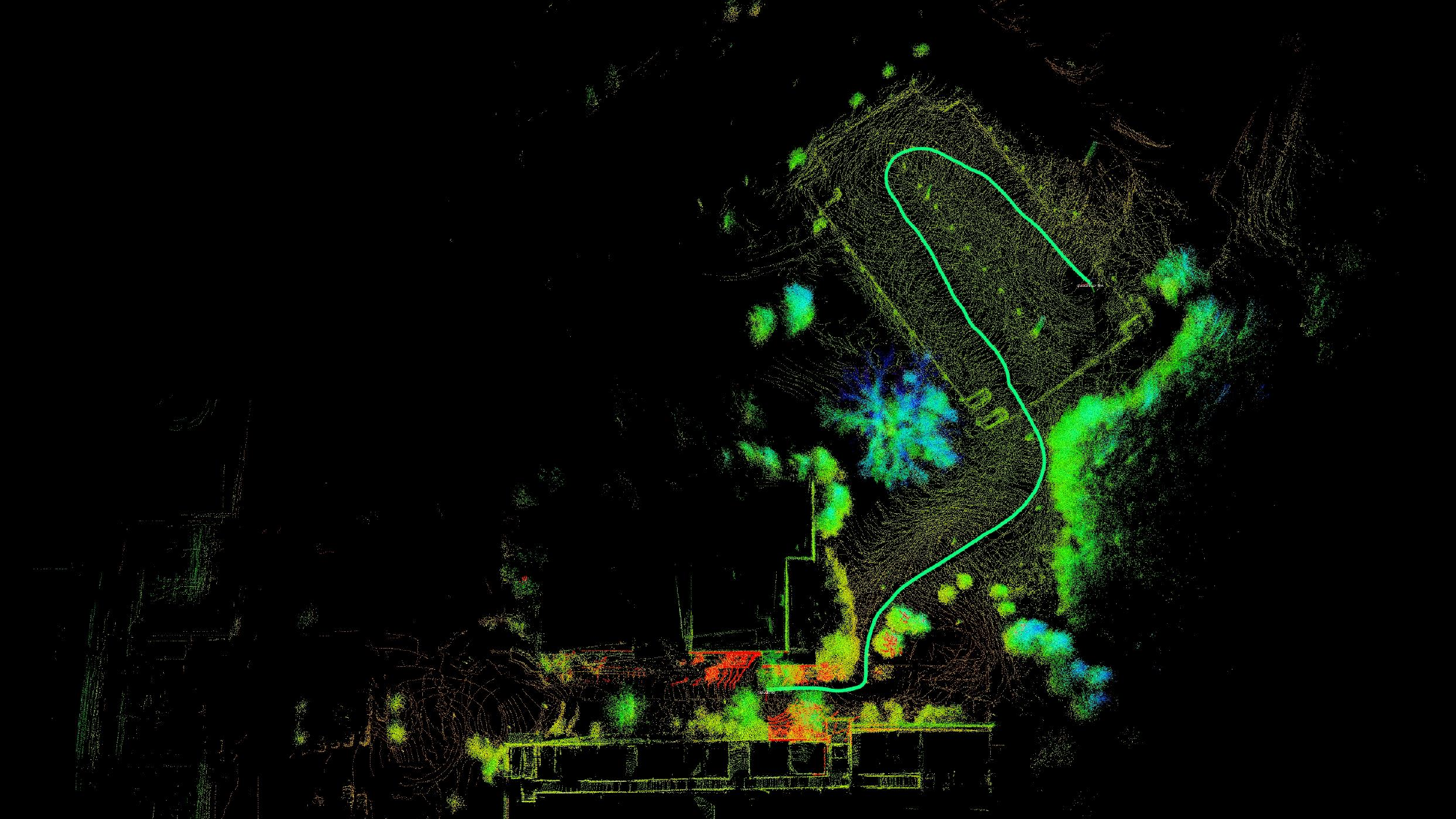}}
    \hfill
    \subfloat[]{\includegraphics[height=0.135\linewidth]{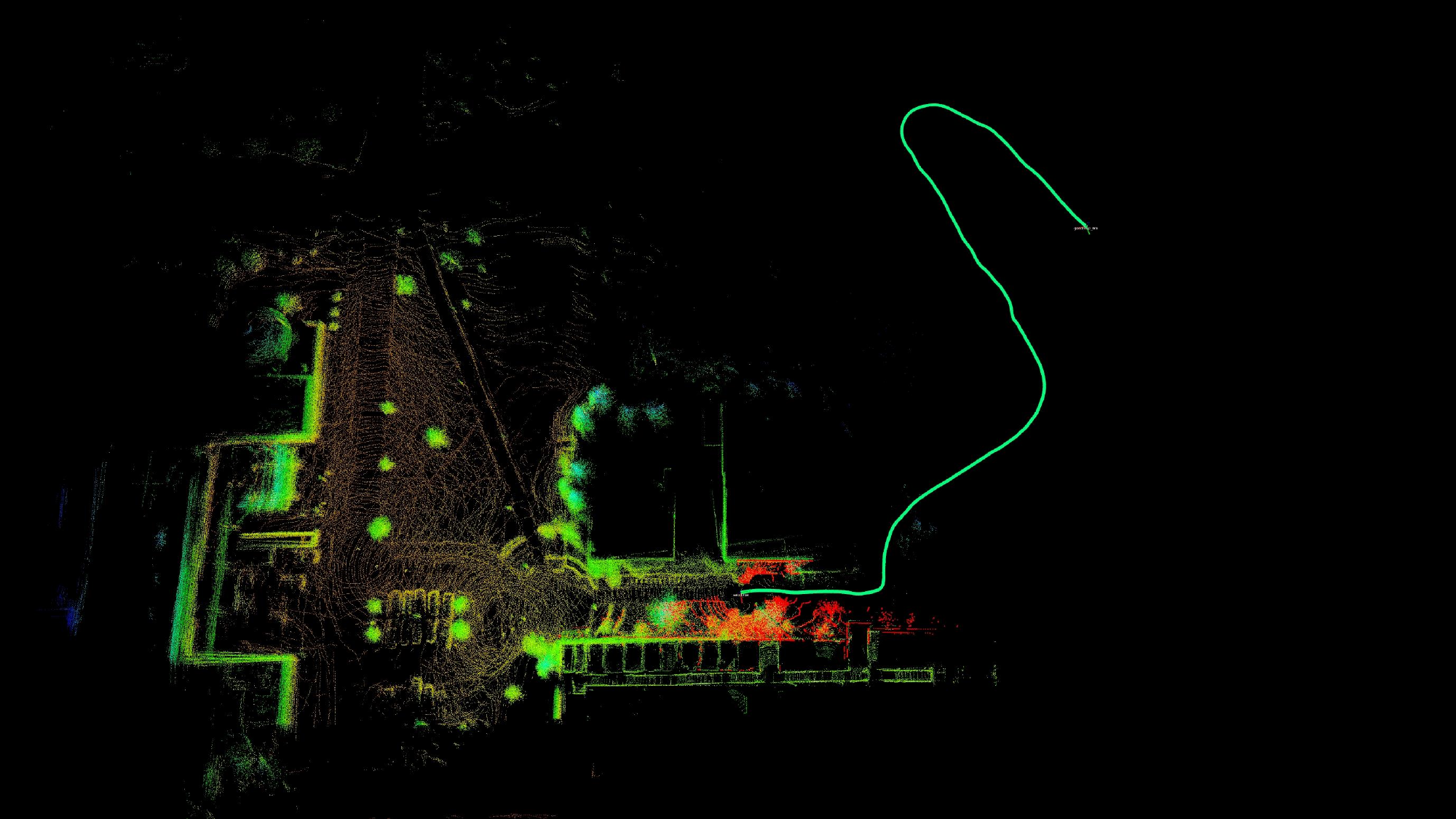}}
    \hfill
    \subfloat[]{\includegraphics[height=0.135\linewidth]{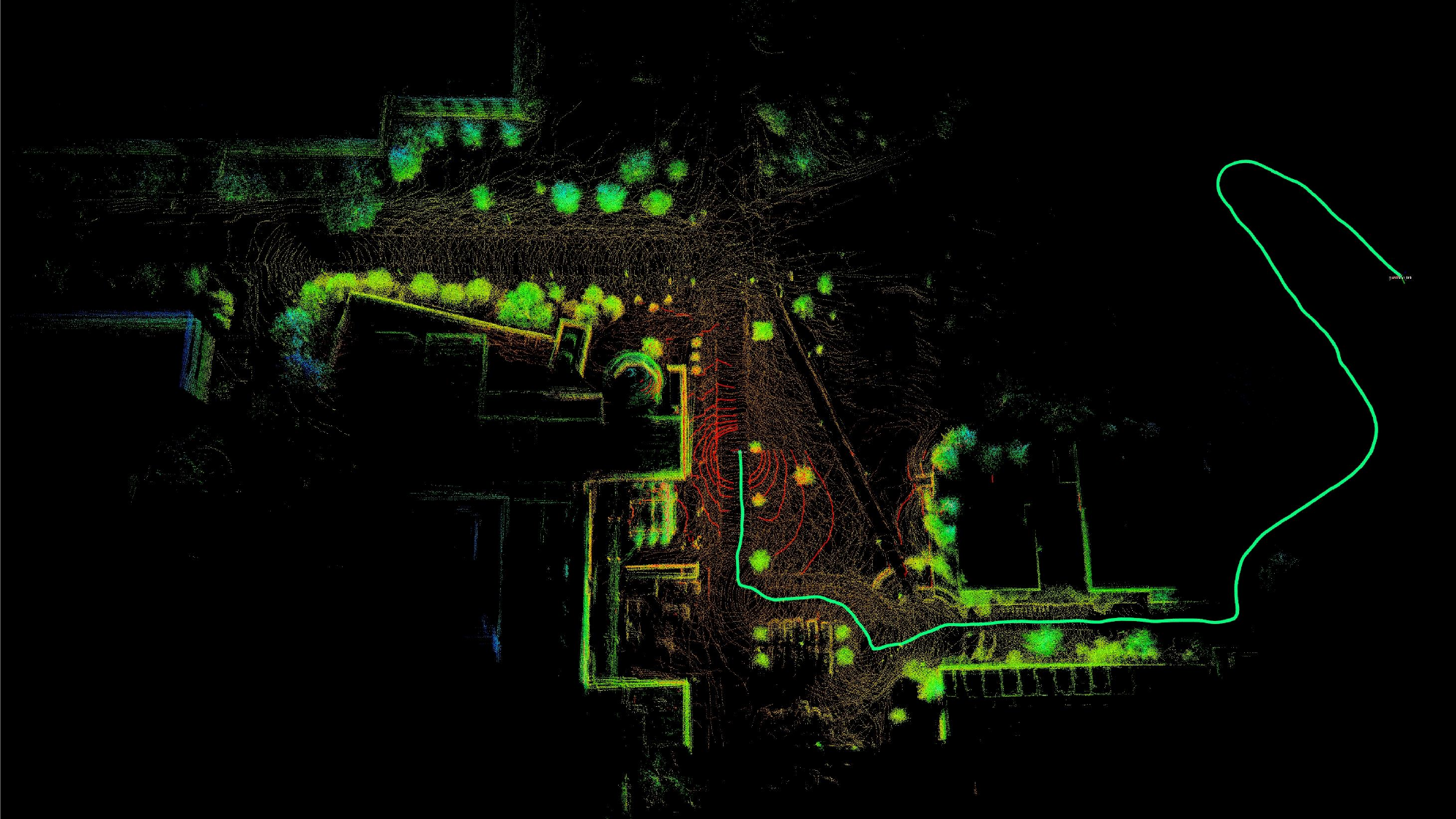}}
    \hfill
    \subfloat[]{\includegraphics[height=0.135\linewidth]{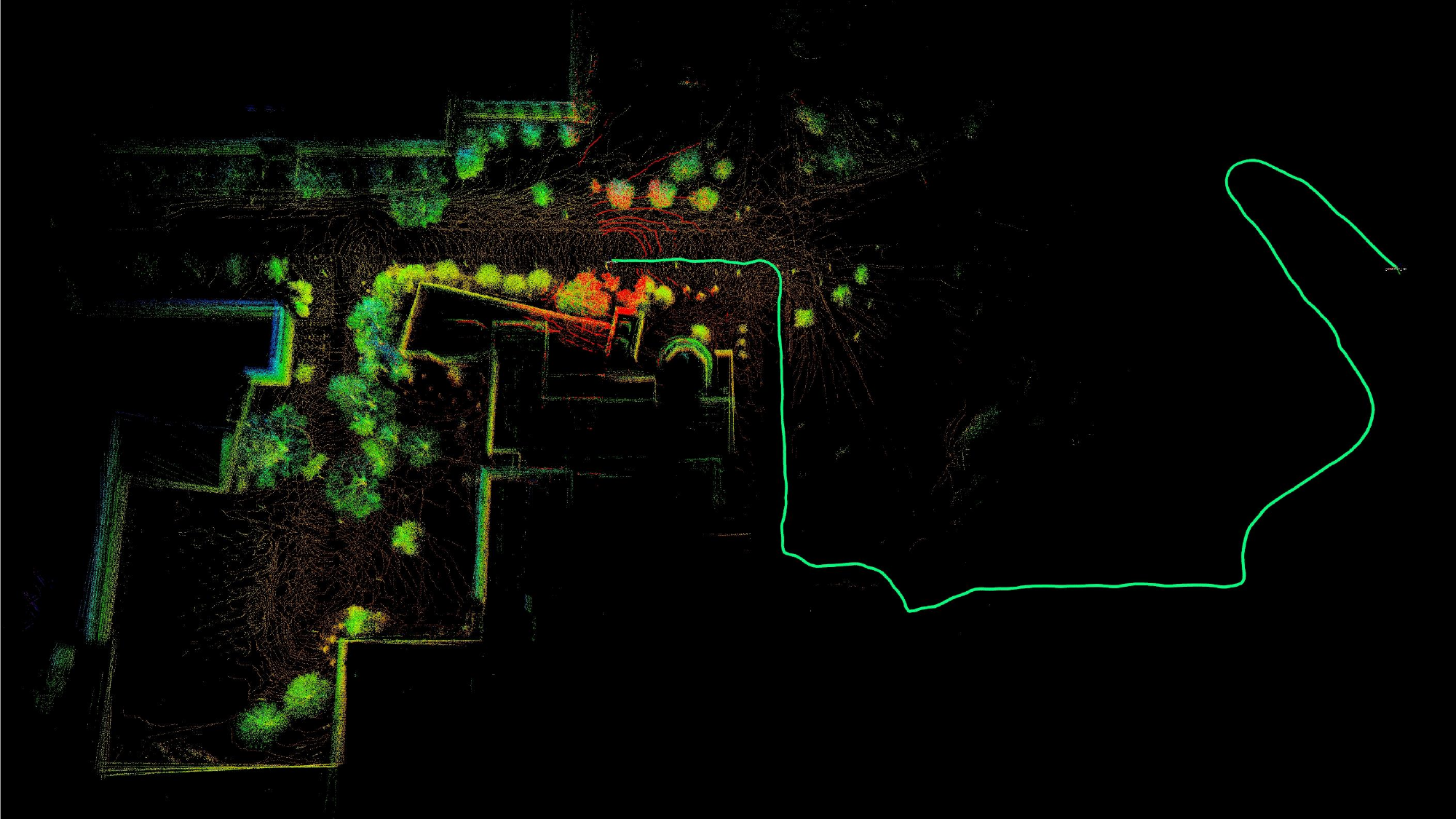}}
    \caption{Subfigures (a) to (d) show smooth transition and the localization path passing through 4 consecutive block maps on \textit{nclt\_2}. Our switching strategy is able to maintain spatial continuity with good performance.}
    \label{fig:switching}
\end{figure*}

\section{Experiment and Results}
\subsection{Dataset}

We evaluate our method on two publicly available datasets, NCLT\cite{nclt} and M2DGR\cite{9664374}. To demonstrate our method's superior performance in large-scale environments, we select the longest sequence from these two datasets, as shown in Table \ref{tab:details}.
Specifically, the North Campus Long-Term (NCLT) dataset, collected at the University of Michigan's North Campus, is a large-scale, long-term autonomy dataset for unmanned ground vehicles. It includes a 10 Hz Velodyne HDL-32E LiDAR and a 100 Hz Microstrain 3DM-GX3-45 IMU for the LIO system. While the same area is repeatedly explored, each sequence features diverse paths and encompasses both outdoor and indoor scenes. We choose five sequences from the NCLT dataset, with two of them extending beyond 6 kilometers in length.

The M2DGR dataset comprises large-scale sequences for ground robots, equipped with a comprehensive sensor suite, including a Velodyne VLP-32C LiDAR and a 150 Hz 9-axis IMU for the LIO system. It records trajectories in challenging situations commonly encountered in practical applications, such as entering elevators and navigating in complete darkness. We select the two longest sequences from the M2DGR dataset to validate the universality of our approach.


\subsection{Pose Initialization}
To evaluate the accuracy of initial pose estimation, we measured the translation
error and the time cost of algorithms compared with various methods including \cite{FPFH,carto,teaser}. 
Given that both \cite{FPFH} and \cite{teaser} consume considerable time initializing map descriptors using the FPFH process (exceeding one hour), rendering them unsuitable for robot global localization. Our approach, compared to the traditional Branch and Bound Search (BBS), employs a greedy strategy to reduce potentially ineffective searches, thereby accelerating localization efficiency. As delineated in Table \ref{tab:gl}, within a search range of 40$\times$40 meters, our method achieves localization times of less than 1 second, ensuring computational precision while striking a balance between computational efficiency and accuracy.
\begin{table}[H]
\renewcommand\arraystretch{1.2}
\centering
\caption{Global localization evaluation}
\label{tab:gl}
\begin{tabular}{lcc}
\hline
  \begin{tabular}[c]{@{}c@{}}Method\end{tabular} &
  \begin{tabular}[c]{@{}c@{}}Translational\\ Error (\textit{m})\end{tabular} &
  \begin{tabular}[c]{@{}c@{}}Time (\textit{s})\end{tabular}
  \\ \hline
 FPFH+RANSAC\cite{FPFH} & 0.26 & 4036.36 \\
                     FPFH+Teaser\cite{teaser} & 0.16 & 3935.50 \\
                    BBS\cite{carto} & 5.00 & 9.82 \\
                    Ours & \textbf{0.05} & \textbf{0.87} \\ \hline
\end{tabular}
\end{table}

\subsection{Pose Tracking Evaluation}
In this section, we implement all experiments on an Intel i9-12700K CPU with 64GB of RAM for fairness. We conducted a rigorous evaluation of two state-of-the-art (SOTA) incremental SLAM-based state estimation algorithms \cite{liosam,xu2022fast} and two map-based localization algorithms Hdl-Localization \cite{hdl} and Fastlio-Loc, which is adapted from FAST-LIO2, loaded the global map once and then utilize the Iterated Extended Kalman Filter (IEKF) for state estimation.
Additionally, to ensure a consistent and equitable comparison with Hdl-Localization, and taking into account the pertinence of control variables, we uniformly employed the NDT registration algorithm predicated on ``DIRECT1". This particular approach entails searching the 1-neighbor voxel to compute the gradient of the matching score. 
The results of our analysis are delineated in Table \ref{tab:eval}. We evaluate the Root Mean Square Error (RMSE) for two distinct trajectory patterns: the ``xyz" mode (trans\_part) and the ``xyzrpy" mode (full). 
Through an in-depth quantitative evaluation, it is evident that our method outperforms other methods on the long-distance NCLT dataset and consistently demonstrates stable performance across all datasets.
This superior performance and stability can largely be attributed to our deployment of a sliding window based on factor graph optimization for state estimation and block maps generation and transition strategy, which exhibits robustness in both outdoor and indoor scenes.
This strategy circumvents the inherent instability of state estimations stemming from the Markov assumption, which is commonly associated with filter-based methodologies.

\begin{table*}[!t]
    \centering
    \caption{absolute trajectory errors (rmse, meters) comparison of state-of-the-arts}
    \renewcommand\arraystretch{1.0}
    \resizebox{0.95\linewidth}{!}
    {
    \begin{threeparttable}
        \begin{tabular}{lrrrrrrrrrrrrrr}
            \toprule
             \multirow{2}{*}{\begin{tabular}[c]{@{}c@{}}Method\end{tabular}}
   &\multicolumn{2}{c}{\textit{nclt\_1}}
   &\multicolumn{2}{c}{\textit{nclt\_2}}
   &\multicolumn{2}{c}{\textit{nclt\_3}}
   &\multicolumn{2}{c}{\textit{nclt\_4}} 
   &\multicolumn{2}{c}{\textit{nclt\_5}}
   &\multicolumn{2}{c}{\textit{m2dgr\_1}}
   &\multicolumn{2}{c}{\textit{m2dgr\_2}}   \\
    \cmidrule(l){2-3}
    \cmidrule(l){4-5}
    \cmidrule(l){6-7}
    \cmidrule(l){8-9}
    \cmidrule(l){10-11}
    \cmidrule(l){12-13}
    \cmidrule(l){14-15}
  &part &full 
  & part &full 
  & part &full
  & part &full
  & part &full
  & part &full
  & part &full 
  \\
            \midrule
            LIO-SAM&$\times$ &$\times$&$\times$&$\times$ &$\times$& $\times$&2.1779 &3.5690&$\times$ &$\times$ &1.4513 &3.1022 & 6.3035 &6.8771 \\
            FAST-LIO2&1.8562& 3.3829 &1.8400& 3.3740 &1.8456& 3.3816 &1.4103& 3.1604 &2.4028 &3.7110&0.4199&2.7773 &2.8482 &3.9557\\
            \midrule
            Hdl-Loc&$\times$& $\times$ &{$\times$}& $\times$&{$\times$}&$\times$ &$\times$&$\times$ &0.5023 &2.8725&$\times$&$\times$&$\times$&$\times$  \\
            Fastlio-Loc&{$\times$}&$\times$ & 3.0193&4.1371&\textbf{0.1795} &2.8340 &0.1973 &2.8352&$\times$&$\times$&\textbf{0.1457}&\textbf{2.7524}&\textbf{0.1302}&\textbf{2.7513}  \\
            \midrule
            Ours&\textbf{0.9736}&\textbf{2.9904} &\textbf{0.1983}& \textbf{2.8345}&0.1854&\textbf{2.8337} & \textbf{0.1832}&\textbf{2.8336}&\textbf{0.1529}&\textbf{2.8316}& 0.2336&2.7561&0.2206&2.7570\\
            \bottomrule
        \end{tabular}
        \begin{tablenotes}
            \footnotesize
            \item[1] $\times$ denotes that the system totally failed.
        \end{tablenotes}
        \label{tab:eval}
    \end{threeparttable}
    }
\end{table*}

\begin{table*}[!t]
    \centering
    \caption{Average Time consumption of state update when using global maps and block maps respectively \protect (ms)}
    \begin{threeparttable}
    \begin{tabular}{lrrrrrrr}
        \toprule
        Method &\textit{nclt\_1}&\textit{nclt\_2}&\textit{nclt\_3}&\textit{nclt\_4}&\textit{nclt\_5}&\textit{m2dgr\_1}&\textit{m2dgr\_2}\\
        \midrule
        Fastlio-Loc& 10.25*&11.12 &10.57 &12.33&12.45*&14.75&11.68\\
        Hdl-Loc&8.55* &8.75* &17.39* &$\times$&19.67&20.52*&$\times$\\
        \midrule
        Ours w/ Global Map&27.35 &35.92 &31.15 &37.25 &29.11 &24.66 &24.43\\
        Ours w/ Block Map&15.25 &14.28 &18.16 &22.25 &20.61 &14.64 &11.17\\
        \bottomrule

    \end{tabular}
    \label{tab:timecomsumption}
    \begin{tablenotes}
    \footnotesize
        \item[1]  $*$ denotes that the system failed in the middle and we calculate the average time before each failure.
        \item[2]  $\times$ denotes that the system totally failed.
    \end{tablenotes}
    \end{threeparttable}
\end{table*}
    

\subsection{Efficiency Evaluation}
The method of map-blocking proposed by Gao \cite{gaoxiang} lacks a mechanism for maintaining a map library, rendering it unsuitable for our evaluation of localization performance. 
Consequently, our comparison is confined to the efficacy of existing map-based localization techniques relative to our own, specifically when employing both global and block maps. As shown in Table \ref{tab:timecomsumption}, the algorithm we employ with the block map is notably faster compared to the one using the entire global map especially when the map is particularly large. In the sequence \textit{nclt\_2}, we have increased the speed by 150\%. However, it is worth noting that as the map continues to grow, our time consumption advantages will become more conspicuous. Additionally, the algorithm based on filtering exhibits an obvious advantage in terms of speed but lacks stability and accuracy. Our block map algorithm is quite close to theirs in terms of speed and performs more robustly in large-scale environments.

\begin{table}[h!]
    \centering
    \caption{details of all the sequences}
    \begin{tabular}{llrr}
        \toprule
        Abbreviation & Name &
        \makecell[c]{Duration \\ (\textit{min}:\textit{sec})}&
        \makecell[c]{Distance \\ (\textit{km})}\\
        \midrule
        \textit{nclt\_1}&20120115&111:46&4.01\\
        \textit{nclt\_2}&20120122&87:19&6.36\\
        \textit{nclt\_3}&20120202&98:37&6.45\\
        \textit{nclt\_4}&20120429&43:17&1.86\\
        \textit{nclt\_5}&20120511&84:32&3.13\\
        \textit{m2dgr\_1}&street\_01&17:08&0.75\\
        \textit{m2dgr\_2}&street\_02&20:27&1.48\\
        \bottomrule
    \end{tabular}
    \label{tab:details}
\end{table}
\section{Conclusion}
In this paper, we have proposed a localization system based on block maps in large-scale environments. To enhance the robustness of robot localization, we have incorporated a factor graph optimization method based on a dynamic sliding window, effectively leveraging more historical information to mitigate the uncertainties associated with the Markov assumption, aligning with our block map switching strategy. Finally, we validated the superiority of our approach on publicly available large-scale datasets. 
In future work, we intend to further enhance this system. We aim to ensure that the robot can achieve real-time and accurate global localization in any position and in highly dynamic environments within the framework of block maps. Additionally, we plan to detect real-time changes in the environment and update the block maps accordingly, catering to applications such as logistics and food delivery robots.

\addtolength{\textheight}{-1cm}   



\bibliographystyle{plain}  
\bibliography{reference}

\end{document}